Corresponding Author:

Mayank Patwari, Technische Universität München, Boltzmanstraße 3, 85748 Garching bei München, email: mayank.patwari@tum.de

Prof. Roozbeh Naemi, Staffordshire University, College Road, Stoke-on-Trent, ST4 2DE, email: r.naemi@staffs.ac.uk

# Title: A quantitative comparison of plantar soft tissue strainability distribution and homogeneity between ulcerated and non-ulcerated patients using strain elastography

Authors : Mayank Patwari [c] MSc, Panagiotis Chatzistergos [a] PhD, Lakshmi Sundar [b] MD, Nachiappan Chockalingam [a] PhD, Ambadi Ramachandran [b] MD, Roozbeh Naemi [a] PhD

Author affiliations:

[a] Faculty of Health Sciences, Staffordshire University, Stoke on Trent, Staffordshire, UK

[b] AR Diabetes Hospitals, Chennai, Tamil Nadu, India

[c] Computer Aided Medical Procedures, Technische Universität München, Munich, Germany

# Abstract

The primary objective of this study was to develop a method that allows accurate quantification of plantar soft tissue stiffness distribution and homogeneity. The secondary aim of this study is to investigate if the differences in soft tissue stiffness distribution and homogeneity can be detected between ulcerated and non-ulcerated foot. Novel measures of individual pixel stiffness, named as quantitative strainability (QS) and relative strainability (RS) were developed. SE data obtained from 39 (9 with active diabetic foot ulcers) patients with diabetic neuropathy. The patients with active diabetic foot ulcer had wound in parts of the foot other than the first metatarsal head and the heel where the elastography measures were conducted. RS was used to measure changes and gradients in the stiffness distribution of plantar soft tissues in participants with and without active diabetic foot ulcer. The plantar soft tissue homogeneity in superior-inferior direction in the left forefoot was significantly ($p<0.05$) higher in ulcerated group compared to non-ulcerated group. The assessment of homogeneity showed potentials to further explain the nature of the change in tissue that can increase internal stress . This can have implications in assessing the vulnerability to soft tissue damage and ulceration in diabetes.

# Keywords

diabetes, elastography, homogeneity, plantar tissue, quantitative, strainability, strain elastography

# Introduction

The plantar fat pad is one of the many parts of the body that is affected by diabetes. Damage to the fat pad is caused due to hyperglycemia, which leads to damage to the nerves of the foot, ulceration, bacterial infection of the ulcer, and finally amputation of the foot (1). Diabetic feet generally have stiffer plantar soft tissue (2), with the stiffness more pronounced in feet that have progressed to ulceration (3). The process of loading the foot was also shown to be of importance, as diabetic ulceration may alter the structure of the soft tissue in such a way that it would have a significantly higher load on the forefoot (4). While the stiffness of plantar soft tissue is associated with an increase in the stress in the soft tissue, the plantar soft tissue homogeneity, which is dependent on the spatial distribution of stiffness, would also affect the internal stress distribution. This is caused by creating areas of high stress concentration inside the soft tissue, which increases the likelihood of mechanical trauma and soft tissue damage. To the best of our knowledge, the use of soft tissue homogeneity as a quantitative metric has not been applied before. This is because, in order to assess soft tissue homogeneity, an accurate measure of the stiffness map of plantar soft tissue is required. Several studies have used methods such as ultrasound indentation techniques to study the overall stiffness of the plantar soft tissue (5–9).

However, these methods do not allow assessing the regional and local differences in stiffness of the tissue that can have practical implications in identifying the areas of high stress concentration inside the tissue that can potentially lead to tissue breakdown and damage.

While strainability of the plantar soft tissue can be assessed using strain elastography proposed by Matteoli et al. (10), to compare strainability results across subjects equal amount of force is needed to be applied to the tissue by the probe. To rectify such issues, in previous studies we proposed the use of stand-off material in conjunction with the use of ultrasound elastogaphy that allows a quantitative assessment of soft tissue stiffness (9). The stand-off material, with known stiffness and strainability, provides a useful reference to estimate the relative strainability of tissue ( Naemi et al, 2016). The results highlighted a lower heel pad stiffness in ulcerated vs non-ulcerated diabetic neuropathic patients (11) and measuring stiffness using this method showed to improve the prognosis accuracy of predicting foot ulcer in diabetic neuropathic patients (12). Although Naemi et al. (11) clearly demonstrated the effectiveness of their method in measuring average relative stiffness of the soft tissue with respect to the stiffness of the standard stand-off material, the proposed method does not allow measurements of tissue homogeneity.

This is because, during strain elastography, the individual points in the soft tissue could be compressed at different level as a result of uneven distribution of load on standoff-skin interface. This is demonstrated by that fact that the homogenous standoff material shows a non-uniform deformation during compression indicating that relative higher forces were applied at the centre of the image compared to the areas away from the central axis. Therefore, areas close to the centre of the soft tissue that are exposed to higher load consequently deform further and these areas could have been mistakenly recognised as the softer regions compared to the areas that are further away from the central axis.

In order to assess the stiffness distribution of the plantar soft tissue, there is a need for quantifying the individual points' stiffness values that takes into account the amount of force that is applied to the soft tissue ensuring an accurate detection of small variations in stiffness. The primary objective of this study was to develop a method that allows accurate quantification of plantar soft tissue stiffness distribution and homogeneity. The secondary aim of this study is to investigate if the differences in soft tissue stiffness distribution and homogeneity can be detected between ulcerated and non-ulcerated foot (13).

## Materials and Methods

### Participants

For the purposes of this study 39 (F/M 6/32) volunteers with type-2 diabetes and diabetic neuropathy, were recruited from the same foot clinic in a specialist diabetes hospital in Chennai, India, between the dates of 11th and 30th June, 2015. The average ± STDEV for age, duration of diabetes and BMI for the Non-ulcerated group were 57±6 years, 14±5 years and 27.2±4.2 kg/m2 respectively. The corresponding values for the ulcerated group were not significantly different as Age: 56±11 years; 15±7 years and 26.8±4.3 (see Table 1).

Table 1. Demographic data of the patients recruited for the study

|  | Age (years) | Total number (F/M) | DURATION OF DM (Y) | Height (cm) | Body mass (kg) | BMI (kg/m$^2$) |
|---|---|---|---|---|---|---|
| Non Ulcerated | 57 ± 6 | 30(5/25) | 14 ± 5 | 170 ± 9 | 79 ± 14 | 27.2 ± 4.2 |
| Ulcerated | 56 ± 11 | 9(1/8) | 15 ± 7 | 168 ± 7 | 76 ± 16 | 26.8 ± 4.3 |

Nine of these patients 9 (F/M 1/8) had an active ulceration at the time of recruitment and testing. Vibration perception threshold (VPT) was measured at three sites at the hallux, first metatarsal head and the heel using a biothesiometer, and subjects with VPT scores more than 25 volts in all these sites were included in the study. VPT measurements were taken on all 3 sites, on both feet, resulting in 6 measurements per patient. Plantar soft tissue stiffness was measured in first metatarsal head and heel areas. Patients that had undergone amputation or had active plantar ulcers in the first metatarsal head and heel areas were excluded from the study. Ethical approval was granted by the Ethics committee and all volunteers had given full and informed consent.

More specifically, 9 of the volunteers had active ulceration on a site other than the regions of interest, for example on the medial and lateral malleoli, hallux, and lateral aspect of the heel for the left foot, and fifth toe, lateral and posterior aspects of the heel, and plantar second metatarsal head for the right foot.

Six volunteers had ulceration on the left foot, two on the right foot, and two on both feet. The ulcers had been present for between 15 and 35 days. The patients with ulcerated forefeet had a standard half shoe, while those with ulcerated heels had normal sandals with a soft microcellular polymer flat insole. Both ulcerated and non-ulcerated subjects were being administered oral hypoglycemic agents, with some additionally requiring insulin.

## Data collection and processing

Real time strain elastography was performed using a linear ultrasound probe (LA533, Esaote S.p.A, Genoa, Liguria, Italy), with a frequency of 13 MHz and a footprint of 53 Å~ 11 mm, and a stand-off (Sonokit, Sonogel Vertriebs Gmbh, Bad Camberg, Hesse, Germany) which according to the manufacturer had properties similar to human soft tissue (sonic velocity 1405 m/s, absorption 0.09 dB/MHz.mm and reflection: 2.4%). The stand-off contact area was 30 Å~ 66 mm and its thickness was 11 mm.

Data was collected from each participant in a 30 minute session with the subjects in a supine position. Real time elastography images involves manually applying a low frequency low magnitude cyclic loading on the imaged tissue with the ultrasound probe. The load is applied to the plantar soft tissue by the operator using the ultrasound probe. A qualitative map of relative strainability is then created by mapping the shift between frames of different tissues (Naemi et al, 2016). Visual feedback on the quality of the generated elastography images was provided by a performance indicator in the software user interface. The plantar soft tissue was compressed between the probe stand-off and calcaneus for measurements at the heel, whereas the plantar soft tissue was compressed between the probe stand-off and the first metatarsal head for the measurement at the forefoot (See Fig. 1).

The regions of interest were those compressed between the bone and the probe. Real time elastography provides a qualitative assessment of strainability, making any comparison between subjects extremely difficult. However, the use of a stand-off material with known mechanical properties as reference enables the qualitative assessment of relative stainability (9) which can be used for comparison between participants.

A custom image processing algorithm was developed in order to extract the relevant features from the elastography videos. This algorithm takes a video as input. Three frames of maximum compression are manually selected from each video that entailed load/unload cycles, from the beginning, middle and end when possible, and processed as images. The bone was automatically removed and the stand-off section is automatically identified and segmented from the soft tissues. The colour of each individual pixel of soft tissue and stand-off is analysed and translated into a strainability score based on the colour code which is visible on the screen. The colour codes indicate the measured strainability i.e. blue and green represent higher strainability and softer tissue, while red and yellow indicate lower strainiability and stiffer tissue. From this point on this strainability score will be referred to as quantitative strainability (QS) and take values between 1 and 100. For the details of the calculation, please see the Appendix section For the details of the calculation, please see the Appendix section.

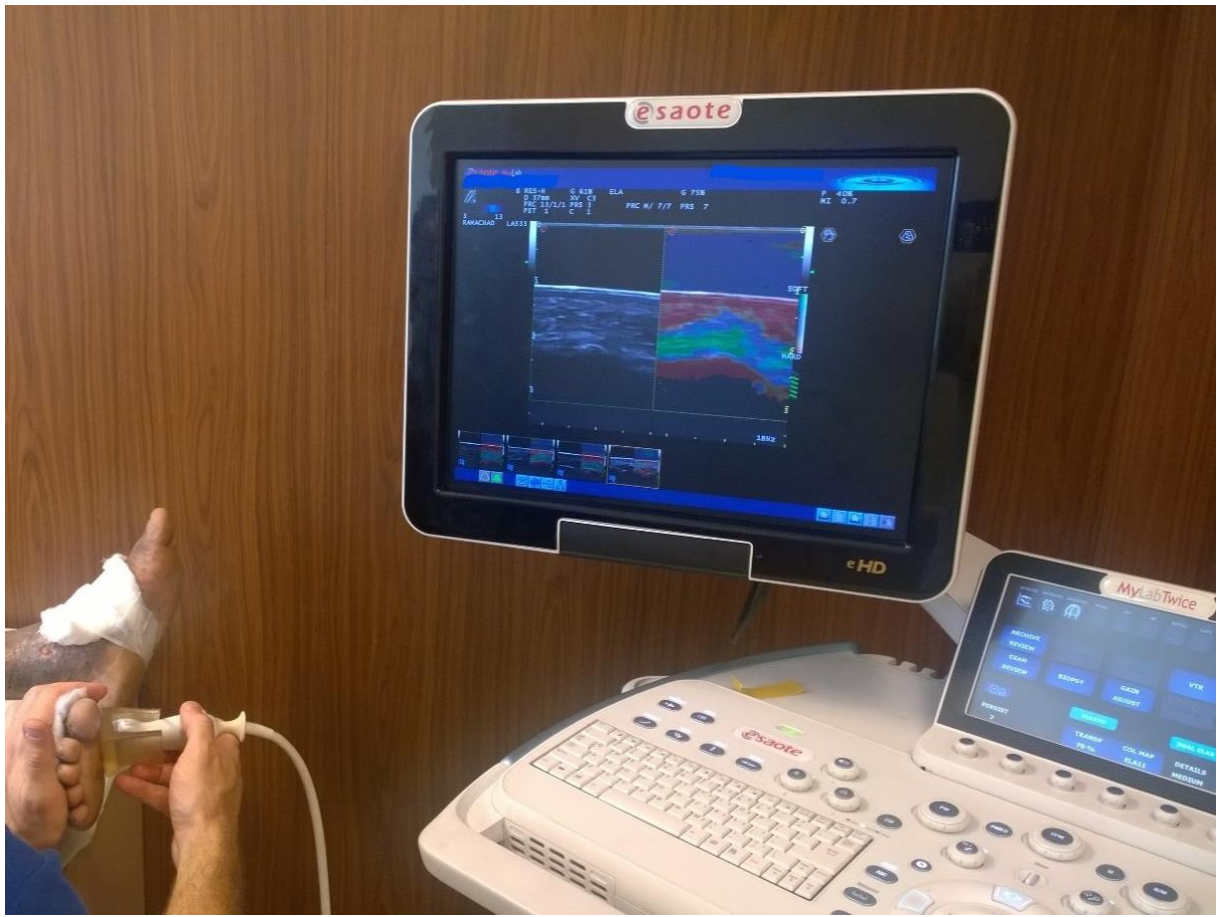

Figure 1 A demonstration of the experimental setup. The device, display, and the participant are all visible

The QS values for all pixels of each column of the stand-off are averaged and a single reference strainability score was calculated for each column of the tissue. This reference strainability score is then used to normalise the QS data of each pixel of the soft tissue based on the average QS of the column of the standoff directly above it, and translate them into relative strainability (RS) scores that can be used for comparison between patients. The RS scores can be used in order to observe the difference in stiffness in the microchambers of the foot (See Fig. 2). According to the strain elastograophy technique proposed by the manufacturer, the force applied by the operator to the tissue needed to be small compression and retraction movements using the probe. Also the forces were perpendicular pressure through rhythmic movements on the tissue under exam, hence each column is considered to be only a compression of the tissue with no shearing and bulging.

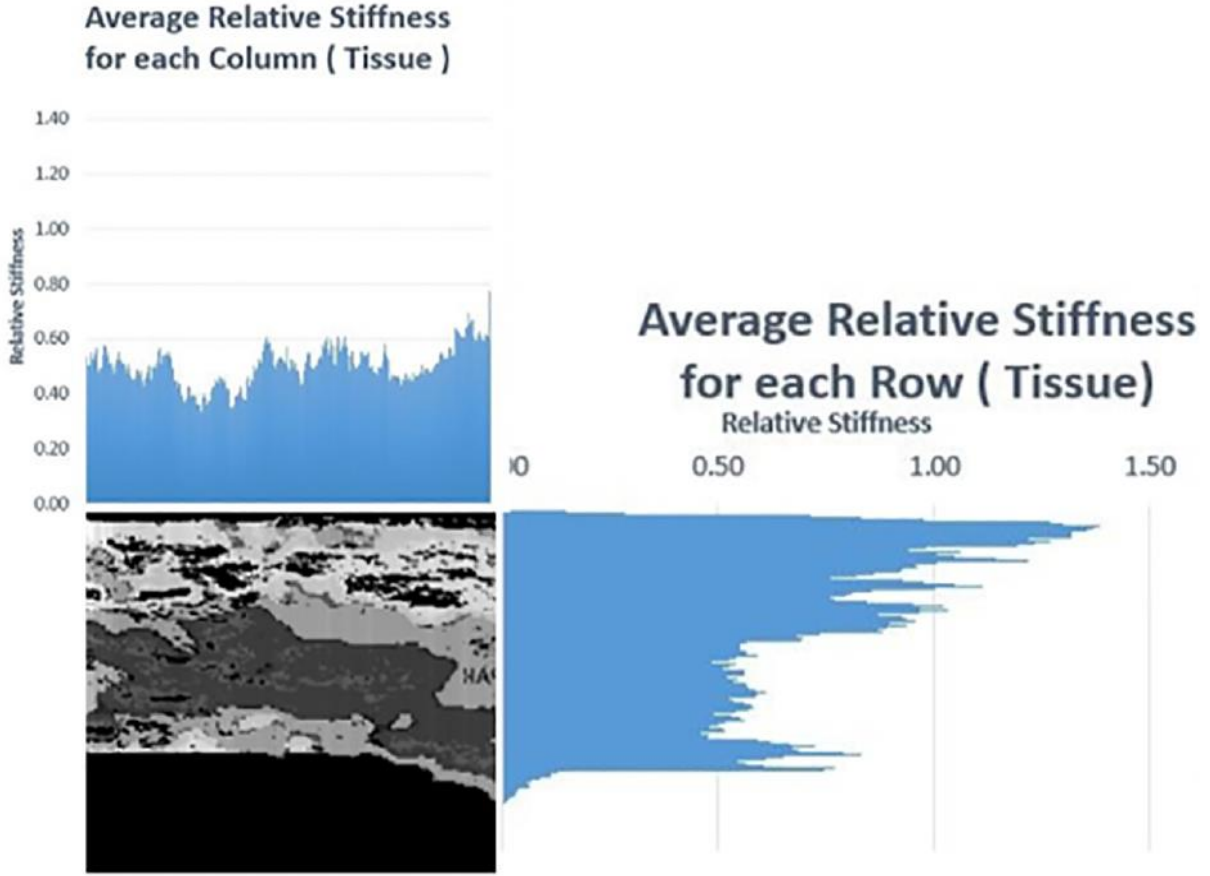

Figure 2 Distribution of row and column stiffness. The row relative stiffness help in outlining the location of the microchambers

The gradients of the RS are extracted in horizontal, vertical and oblique directions using the right difference methods. The gradients in the horizontal direction are referred to by $G_x$, those in the vertical direction as $G_y$, and the oblique gradient by $G_r$ (See Fig. 3). The three gradient vectors are calculated by the following equations:

$$G_{xi} = \frac{RS_i - RS_{i-1}}{x_i - x_{i-1}} \quad \text{(Equation 1)}$$

$$G_{yi} = \frac{RS_i - RS_{i-1}}{y_i - y_{i-1}} \quad \text{(Equation 2)}$$

$$G_{ri} = \sqrt{G_{xi}^2 + G_{yi}^2} \quad \text{(Equation 3)}$$

where $G_{xi}$, $G_{yi}$ and $G_{ri}$ represents stiffness gradient for cell i along the breadth, depth and oblique directions respectively. The mean gradient value of each row for $G_x$ and the mean gradient for each column for $G_y$ are calculated in order to measure changes (See Fig. 4).

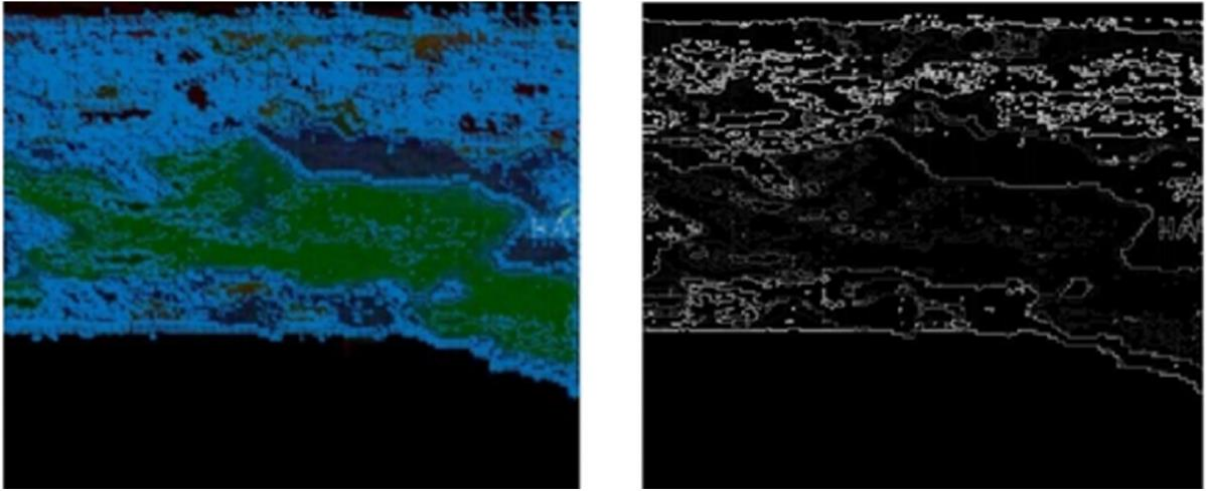

Figure 3 Map of oblique gradients in an image

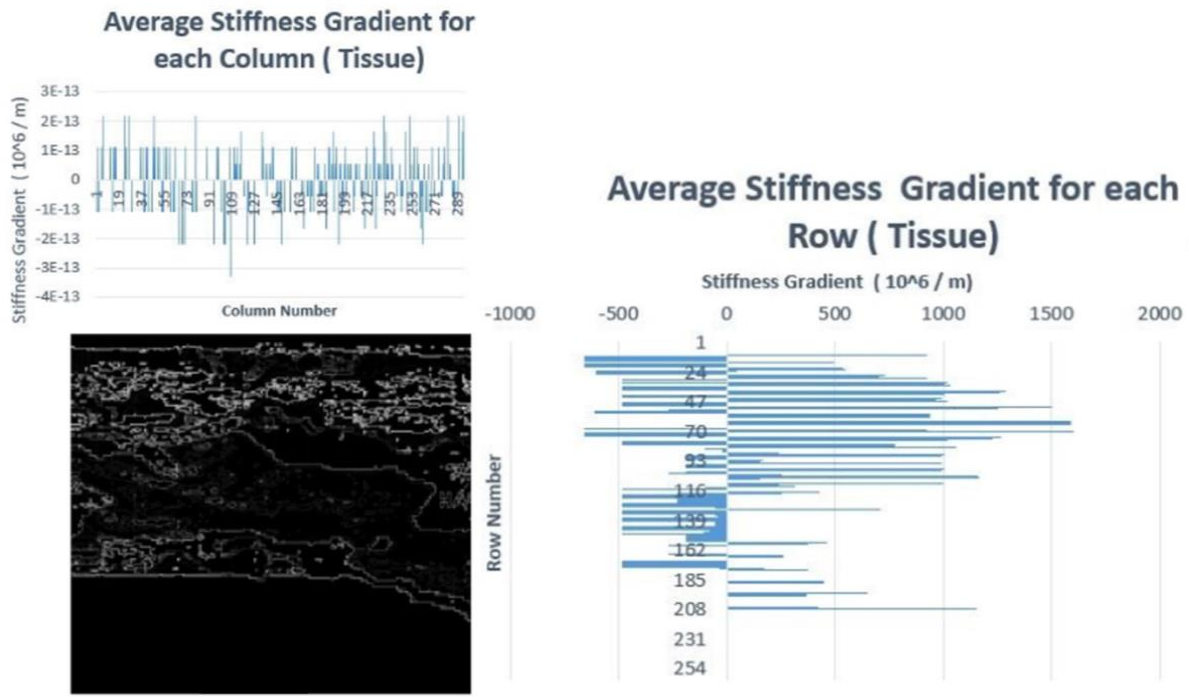

Figure 4 Average of stiffness gradient by row and column. The average of Gx for each row is shown on the right, whereas the average for Gy for each column is shown above the image

A positive anterior-posterior gradient indicates stiffening towards the posterior direction, while a negative gradient indicates stiffening towards the anterior direction. Similarly, a positive superior-inferior gradient indicates stiffening towards the bone in the superior direction, while a negative superior-inferior gradient indicates stiffening towards the skin surface in the inferior direction.

The total gradient in each direction is calculated as the mean gradient of the mean gradients for each row (Gx) and each column (Gy). The total oblique gradient is calculated by using Equation 3 with the total X gradient and total Y gradient.

Four different cases are considered when analysing the results:

- Left forefoot
- Left heel
- Right forefoot
- Right heel

Three frames are extracted at random for each video, for the left and right forefeet and heels, resulting in a total of 12 frames to analyse for each subject. Due to inherent differences between left and right foot of each participant and their dependencies (Naemi et al, 2016) it was not possible to mix the data from both feet together and all analyses were done separately for the left and the right feet. Each frame was treated as a separate image and the data was treated as belonging to a separate data source.

# Results

The calculated anterior-posterior gradients do not show any statistically significant difference between ulcerated and non-ulcerated feet on either the forefoot or the heel (Fig. 5). The superior-inferior gradient is significantly higher for non-ulcerated left forefoot ($t = 1.992$, $p = 0.049$, $\eta 2 = 0.000$). This indicates that the plantar soft tissue homogeneity in superior-inferior direction at the left forefoot was significantly different between the ulcerated and non-ulcerated feet (Fig. 6). This also implies that ulcerated left forefeet have higher uniformity and homogeneity than non-ulcerated left forefeet The calculated net (oblique) vectors do not show any statistically significant difference between ulcerated and non-ulcerated feet on either the forefoot or the heel (Fig. 7).

# Discussion

## Mechanical Properties

This study looks at the distribution of stiffness gradients in the foot as an attempt to observe a link between plantar soft tissue homogeneity and diabetic foot ulceration. This study observed a link between diabetic ulceration and decreased homogeneity in the superior-inferior direction of the left forefoot.

Matteoli et al. (10) showed the presence of a soft layer toward the bone, and a harder layer towards the skin, meaning that in all cases there would be a stiffening toward the surface of the skin, with a highly negative superior-inferior gradient. Non-ulcerated feet however, show a highly positive superior-inferior gradient, which indicates stiffening away from the skin and in the direction of bone. This could be explained by the way the tissues are distributed in the foot. An example can be seen in Fig. 3, where there are small gradients in the general direction within the green areas, but large gradients in the opposite direction in the interfaces between red and blue areas, and again between blue and green areas. The average calculated gradient could indeed be positive.

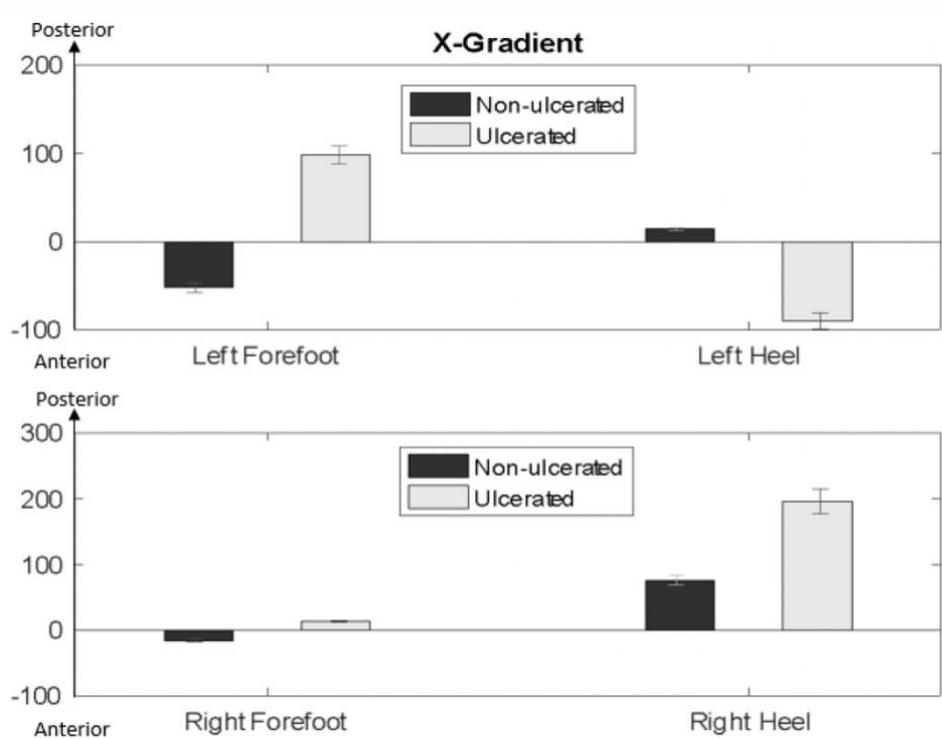

Figure 5 Graph comparing Anterior-posterior gradients at four different positions – the left and right heels, and the left and right forefeet near the 1st metahead. This are present with 95% confidence bars, and all significant differences are marked with *

Ulcerated feet show negligible superior-inferior gradients. There is almost no variance between the layers, which may indicate that the plantar fat pad has been worn down to such an extent that the difference between the microchamber layers and microchamber layers is almost indistinguishable. This trend is present in both the forefoot and the heel, which suggests uniform deterioration of fat pad due to ulceration. An example can be seen in Fig. 8, where the green layers no longer show visibly different stiffnesses.

The anterior-posterior gradients in non-ulcerated feet show uniform trends, with only mild stiffening in the anterior and posterior direction. However, the anterior-posterior gradients on ulcerated feet are on average quite large, showing that there is stiffening in a particular direction, towards the posterior side for the left forefoot and right heel, and towards the anterior side for the left heel. The right forefoot does not have strong gradients. Since there is a noted higher stiffness in the heels, an accumulation of stiffness gradients toward the heels (located on the posterior) would be expected. However, the opposite trend is observed. A possible reason for stiffening toward the anterior direction is that the posterior part of the heel pad is the first point of contact with the ground during a foot strike. This could have resulted in an increased stiffness in this area to reinforce structures to endure larger forces (14).

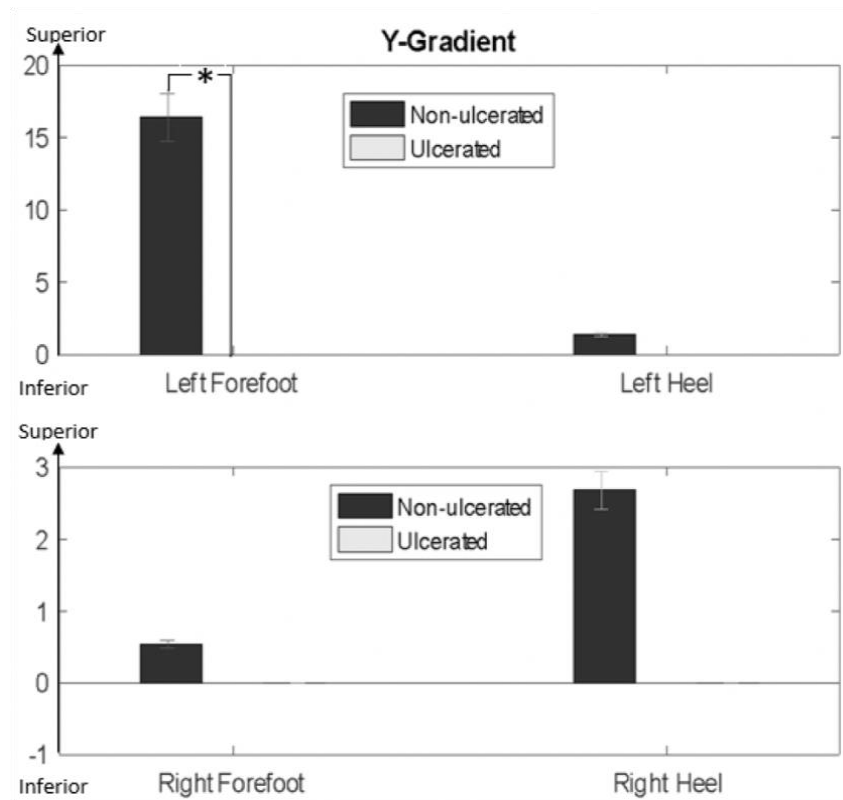

Figure 6 Graph comparing Superior-inferior gradients at four different positions – the left and right heels, and the left and right forefeet near the 1st metahead. This are present with 95% confidence bars, and all significant differences are marked with *

The oblique gradients are relatively uniform for both ulcerated and non-ulcerated diabetic feet. The oblique gradients are calculated by taking the vector sum of both the superior-inferior and anterior-posterior gradients. Since the gradient results are quite contrasting for the ulcerated vs non-ulcerated in the two types of feet, results in these components complementing each other when the oblique gradient is calculated.

While the net vector provides an overall measure of homogeneity in both direction, the finding of this study where there superior-inferior homogeneity was found to be significantly lower in ulcerated patients can have huge practical implications. This can indicate a different gradient of stainability of the tissue that lead to an increased internal stress in the soft tissue that can be linked back to enhanced risk of tissue damage and ulceration.

Homogeneity in each direction can indicate the internal stress concentration/distribution hence the vulnerability to tissue damage in different directions The superior/inferior and anterior-posterior direction homogeneities can be linked to the vulnerabilities to tissue damage in these directions as a result of loading in the vertical and horizontal (direction of walking) direction during stance phase of gait. In addition the homogeneity in oblique direction can be linked to the vulnerability in the resultant forces that applies to the plantar soft tissue as a result of both vertical and anterio-posterior components of the ground reaction force during walking.

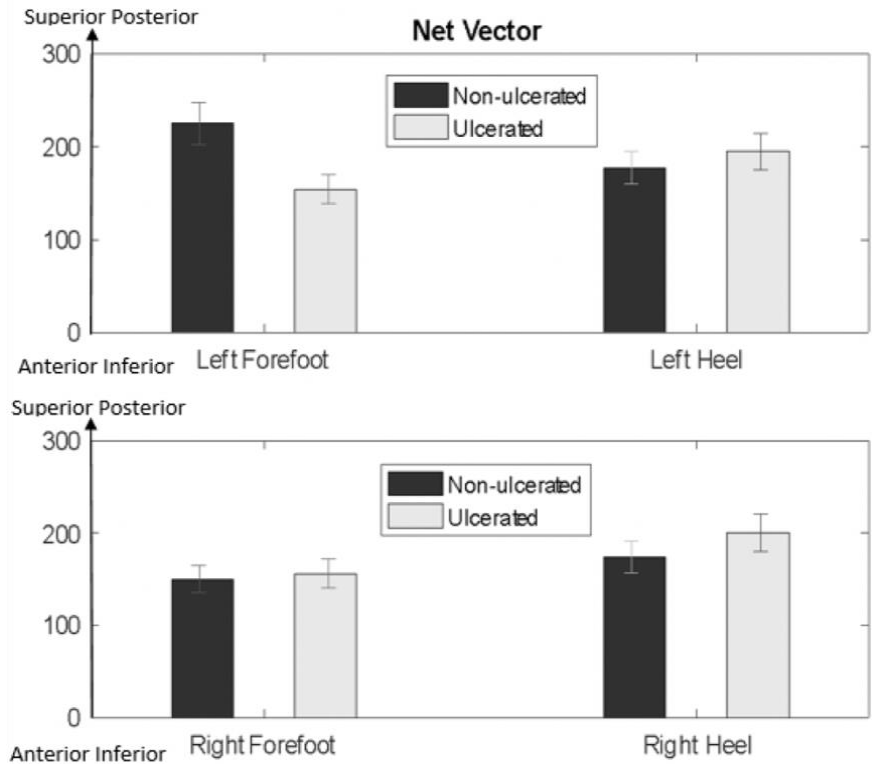

Figure 7 Graph showing net stiffness vectors at four different positions - the left and right heels, and the left and right forefeet near the 1st metahead. This are present with 95% confidence bars, and all significant differences are marked with *

## Relation to other studies

The use of pixel strainability in this study allowed for each individual section of the image to be analysed separately, which enabled the calculation of stiffness gradients. The two sections of harder and softer tissue as shown by Matteoli et al. (10) can be clearly differentiated, with the harder section being red and closer to the skin, and the softer sections being blue and green, and closer to the bone. Strain elastography has already been used by Naemi et al.(11) to differentiate between the ulcerated and non-ulcerated feet in terms of the mechanical properties of the plantar soft tissue and by Deprez et al. (15) to detect ulceration by observing changes in each of the structures. This study has shown that quantitative strain elastography can be used as a method to measure homogeneity. We propose that quantitative homogeneity can be used to assess the aetiology and the effect of ulceration.

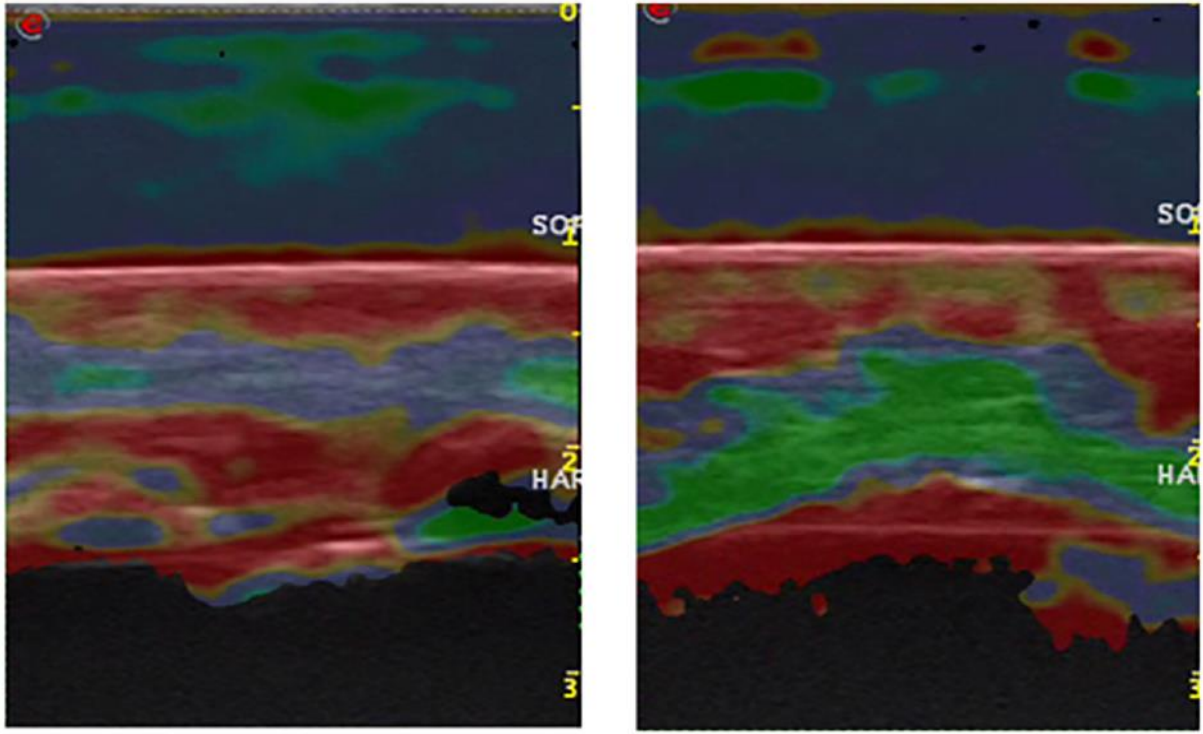

Figure 8 Image of heel in an ulcerated foot (left) vs non-ulcerated foot (right). The erosion of the green layers in the ulcerated foot are clearly visible in the left image, as opposed to the clearly separated layers in the right image

## Clinical significance

The use of foot homogeneity and stiffness gradients as a measure of the soft tissue status has not been previously conducted in the area of diabetic foot disease. While previous studies have shown the differences in plantar soft tissue stiffness between ulcerated and non-ulcerated foot (14), difference in plantar soft tissue homogeneity has never been established before. The current study for the first time indicates that significant differences in the tissue homogeneity in superior-inferior direction could exist between ulcerated and non-ulcerated foot. While at this stage it cannot be concluded whether the observed difference in the homogeneity is a cause or consequence of the ulceration, the difference can indicate association with mechanical trauma and ulceration. This can be used to assess the sequence of events that lead to ulceration and may be used as an early warning system for diagnosing sign of tissue deterioration. This would reduce the costs associated with treatment and decrease the rate of amputations.

## Limitations of the method

Strain from the transducer can only be applied in the vertical (superior-inferior) direction. This could be a reason why a significant difference in homogeneity was observed only in the superior-inferior direction. Accurate measurement of homogeneity in the horizontal (anterior-posterior) direction would not be possible with quantitative strain elastography. Measuring the strainability in the horizontal direction would be possible with the use of the lateral forces applied by shear wave elastography. Furthermore, while the knowledge about the homogeneity can have implications in early diagnosis of soft tissue damage in diabetic patients, further prospective studies are required to investigate the role of tissue homogeneity in predicting diabetic foot ulceration. This can also help to determine whether the differences observed is physiological changes that contribute to ulceration or pathophysiological changes that happen after ulceration as a result of altered loading or other pathophysiological phenomena. It needs to be mentioned that because of a lower number of ulcerations in the right feet the results were not significant for the right foot. Also due to inherent differences between left and right foot of each participant and their dependencies it was not possible to mix the data from both feet together. This warrant further studies in which a bigger cohort of patients can be studied where the number of ulcers in each foot is higher.

# Summary

This study proposes homogeneity as a measure of diabetic ulceration. Homogeneity in the anterior superior direction shows significant differences between ulcerated and nonulcerated feet. An accurate calculation of homogeneity requires accurate information about soft tissue strainability. The quantitative methods described in this paper can be used to accurately extract soft tissue strainability values and calculate homogeneity. This can have implications in diagnosing soft tissue status and assess its vulnerability to soft tissue damage and ulceration.

# Acknowledgements

Technical support from Esaote S.p.A is acknowledged.

Funding Source: This project led by Staffordshire University is funded by the European

Commission through Grant Agreement Number 285985 under Industry Academia partnerships and Pathways (FP7-PEOPLE-2011-IAPP)

# Appendix

## Quantitative strain elastography methodology detailed

The post-processing algorithm uses the video of strain elastography as an input. Each frame of the input video is processed into a single image. Each image contains 3 parts: the standoff interface, the soft tissue, and the bone. The aim of the algorithm is to extract and normalise the data from the soft tissue. Removal of extraneous information is performed by subtracting the standard B-mode ultrasound image from the elastography image. This removes image data which may interfere with the analysis of the stiffness data. Bone data is removed by looking for pixels with high QS that are distant from the skin i.e. high stiffness values in the bottom parts of the image, and removing them. After removal of ultrasound image data, the skin is represented as a thin line in the stiffness image with no stiffness values present. This enables the easy extraction of the standoff as all the stiffness values above the skin interface. The standoff can then easily be removed from the image, leaving only the soft tissue data. The standoff data will however be retained for normalisation. A scale for mapping colours in the soft tissue and standoff data is developed from the reference colour bar present in the image. This scale maps colour intensities to values from 1-100, with the absence of stiffness data represented as 0. Red coloured values are higher in stiffness, while green coloured values are lower in stiffness. This scale is applied to the soft tissue data as well as the standoff data, in order to get the stiffness value in each pixel. Normalisation occurs by finding the average stiffness value for each column of pixels in the standoff. Each pixel in the corresponding column in the soft tissue data is divided by the average stiffness value of the standoff in that particular column. This normalised data can then be processed to return stiffness values, homogeneity or gradient values, and other calculations. An example of this being applied to retrieve homogeneity values for the diabetic foot is shown in this study.